\documentclass[11pt, a4paper, logo]{deepmind}
\usepackage{kantlipsum, lipsum}
\usepackage{dsfont}
\usepackage[authoryear, sort&compress, round]{natbib}
\usepackage{subfigure}
\usepackage{wrapfig}

\newcommand{\cR}{\mathcal{R}}

\newcommand{\cX}{\mathcal{X}}
\newcommand{\cY}{\mathcal{Y}}

\newcommand{\RaterTrain}{\emph{CleanLabelTrain}}
\newcommand{\SynthTrain}{\emph{NoisyLabelTrain}}
\newcommand{\RaterValid}{\emph{CleanLabelValid}}
\newcommand{\SynthValid}{\emph{NoisyLabelValid}}
\newcommand{\Test}{\emph{Test}}
\newcommand{\kalpha}{{\fontfamily{phv}\selectfont{k-alpha}}}
\newcommand{\LQM}{\mathsf{LQM}}
\graphicspath{{figures/}}

\title{An Instance-Dependent Simulation Framework for Learning with Label Noise}

\reportnumber{} % Leave blank if n/a 
\author[1]{Keren Gu}
\author[1]{Xander Masotto}
\author[1]{Vandana Bachani}
\author[2]{Balaji Lakshminarayanan}
\author[1]{Jack Nikodem}
\author[1]{Dong Yin}
\affil[1]{DeepMind}
\affil[2]{Google Research, Brain Team,
Emails: \{kerengu, xanderm, vbachani, balajiln, nikodem, dongyin\}@google.com}

\begin{abstract}
We propose a simulation framework for generating instance-dependent noisy labels via a pseudo-labeling paradigm.
We show that the distribution of the synthetic noisy labels generated with our framework is closer to human labels compared to independent and class-conditional random flipping.
Equipped with controllable label noise, we study the negative impact of noisy labels across a few practical settings to understand when label noise is more problematic.
We also benchmark several existing algorithms for learning with noisy labels and compare their behavior on our synthetic datasets and on the datasets with independent random label noise.
Additionally, with the availability of annotator information from our simulation framework, we propose a new technique, Label Quality Model (LQM), that leverages annotator features to predict and correct against noisy labels. We show that by adding LQM as a label correction step before applying existing noisy label techniques, we can further improve the models' performance. The synthetic datasets that we generated in this work are released at \url{https://github.com/deepmind/deepmind-research/tree/master/noisy_label}.
\end{abstract}

\begin{document}

\maketitle

\section{Introduction}\label{sec:intro}
In many applications, training machine learning models requires labeled data. In practice, the training data labeled by human raters are often noisy, leading to inferior model performance. The study of learning in the presence of label noise dates back to the eighties \citep{angluin1988learning}, and still receives significant attention in recent years \citep{natarajan2013learning,reed2014training,malach2017decoupling,han2018co,li2020dividemix}.

In the research community, some datasets with real noisy human ratings are available, such as Clothing 1M \citep{xiao2015learning}, Food 101-N \citep{lee2018cleannet} (only a small subset has clean labels), WebVision~\citep{li2017webvision}, and CivilComments \citep{borkan2019nuanced}, which allow testing approaches that address label noise. However, since the level and type of label noise in these datesets cannot be controlled, it becomes hard to conduct ablation study to understand the impact of noisy labels.
As a result, the majority of research work in this area uses benchmark datasets generated by simulations. For example, many prior works simulate noisy labels by flipping the labels according to certain transition matrix \citep{natarajan2013learning,khetan2017learning,patrini2017making,han2018co,hendrycks2018using}, independently from the model inputs, e.g., the raw images. However, this type of random label noise may not be an ideal way to simulate noisy labels, since the errors in human ratings are often instance-dependent, i.e., harder examples are easier to get wrong labels, whereas the noisy labels generated by random flipping do not have this type of dependency, even if the transition matrix is asymmetric, i.e., class-conditional. In addition, in many applications, we often have additional features of the raters, such as tenure, historical biases, and expertise level~\citep{cabitza2020elephant}. Leveraging these features properly can potentially lead to better model performance. However, neither the commonly used public datasets with human ratings nor the synthetic datasets created by random label noise have such rater features available.

In this work, we focus on creating simulation benchmarks for the research on label noise. We propose a method that is instance-dependent, easy to implement, and can convert any commonly used public dataset with clean labels into a noisy label dataset with additional rater features. More specifically, we propose a simulation method based on a pseudo-labeling paradigm: given a dataset with clean labels, we use a subset of it to train a set of models (rater models), and use them to label the rest of the data. In this way, we obtain a dataset whose size is smaller than the original one with clean labels, but with multiple instance-dependent noisy labels. Moreover, some characteristics of the rater models, such as the number of training epochs, the number of samples used, the validation performance metrics, and the number of parameters in the model can be used as a proxy for the rater features.

We note that this simulation approach is very similar to self-training in semi-supervised learning \citep{chapelle2006semi}. In the research on label noise, methods inspired by semi-supervised learning have been adopted in several prior works for training robust models~\citep{han2018co,li2020dividemix} or generate synthetic noisy label dataset~\citep{lee2019robust,robinson2020strength}. We intend to exploit this approach for both providing a comprehensive study of how practical label noise affects the performance of machine learning models, and the research of better training algorithm in the presence of label noise. Our main contributions are summarized as follows:
\begin{itemize}\setlength\itemsep{0.3em}
    \item We propose a pseudo-labeling simulation framework for learning with label noise. We provide detailed description, including the generation of rater features. We also evaluate the synthetic datasets generated by our framework and show that the distribution of noise labels in our datasets is closer to human labels compared to independent and class-conditional random flipping (Section~\ref{sec:generation_all}).
    
    \item We study the negative impact of label noise on deep learning models using our synthetic datasets. We find that noisy labels are more detrimental under class imbalanced settings, when pretraining is not used, and on tasks that are easier to learn with clean labels (Section~\ref{sec:impact_analysis}).
    
    \item We benchmark existing approaches to tackling label noise using our synthetic datasets. We find that the behavior of these techniques on the our synthetic datasets is different from the datasets generated by independent random label flipping. With the same fraction of mislabeled data, our datasets tend to be harder than datasets with random label noise for binary classification tasks; however, we observe the opposite trend for multi-class tasks (Section~\ref{sec:benchmark}).
    
    \item We propose a label correction approach, named Label Quality Model (LQM), that leverages rater features to significantly improve model performance. We also show that LQM can be combined with other existing noisy label techniques to further improve the performance (Section~\ref{sec:lqm}).
\end{itemize}

\section{Generating Synthetic Datasets with Instance-Dependent Label Noise}\label{sec:generation_all}

In this section, we discuss the formulation of generating synthetic noisy labels, and provide details for the dataset generation procedure and the methods we use to evaluate whether the synthetic datasets share certain characteristics of real human labeled data.

\subsection{Formulation}\label{sec:formulation}

We consider a $K$-class classification problem with input space $\cX$ and label space $\cY=\{1,\ldots, K\}$. In addition, we assume that there is a rater space $\cR$, with each element being the feature of a rater who can label any element in $\cX$. Suppose that there is an unknown distribution over $\cX \times \cY \times \cR \times \cY$, and each tuple $(x, y^*, r, y)$ in this space corresponds to the input feature of an example $x$, clean label of the example $y^*$, a rater $r$, and the label $y$ provided by the rater.

The problem of generating synthetic noisy labels can be modeled as generating a noisy label $y$ given a pair of input feature and clean label $y^*$. Ideally, the probability distribution of the noisy label $y$ should depend on all of $x$, $y^*$, and $r$, i.e., we should generate $y$ according to $p(y\mid x, y^*, r)$. This means that the label noise should depend on the input---harder and more nuanced examples such as blurred images are more likely to have incorrect labels, as well as the rater---raters with higher expertise level are less likely to make mistakes.

However, many prior studies on generating synthetic noisy labels ignore such dependency on $x$ and $r$ and only generate $y$ according to $y^*$. Here, we specify three approaches for generating noisy labels.

\begin{itemize}\setlength\itemsep{0.3em}

    \item \emph{Independent random flipping}. In this method, with probability $\delta$, the label of each example is flipped to an incorrect one, uniformly chosen from all the other $K-1$ labels~\citep{zhang2021understanding,rolnick2017deep,han2018co}. The method is sometimes called symmetric label noise. More specifically, we have
   \[
   p(y=k \mid y^*) = (1-\delta)\mathbf{1}(k=y^*) + \frac{\delta}{K-1}\mathbf{1}(k\neq y^*).
   \]

    \item \emph{Class-conditional random flipping}. In this method, we assume that there is a stochastic matrix $T\in\mathbb{R}^{K\times K}$. The $i$-th row of $T$ corresponds to the probability distribution of the noisy label $y$ given that the clean label $y^*=i$, i.e.,
    \[
    p(y=j \mid y^*=i)=T_{i,j}.
    \]
    This method is sometimes called the asymmetric label noise, and is usually considered more practical than symmetric noise, since classes that are semantically close are more likely to be confused than classes that have clearer decision boundaries. As mentioned in Section~\ref{sec:intro}, this method has been used in many prior works~\citep{angluin1988learning,han2018co,zhang2017mixup,wang2019symmetric,jiang2018mentornet}; the matrix can be designed with human knowledge or estimated from a small subset of clean data~\citep{patrini2017making,hendrycks2018using}. Here, we emphasize that the noisy labels in class-conditional label flipping still do not depend on the input feature $x$ and the rater $r$.
    
    \item \emph{Instance-dependent}, i.e., generating noisy labels according to $p(y \mid x, y^*, r)$. The method that we propose in this paper satisfies this criterion. Similar problems have been considered in several very recent works~\citep{berthon2021confidence,zhu2021second,zhang2021learning}.

\end{itemize}

\subsection{Dataset generation}\label{sec:dataset_gen}
In our framework, we first identify a public dataset that we would like to generate noisy labels for, e.g., CIFAR10 \citep{krizhevsky2009learning} for image classification. We observe that many public datasets already have default training, validation, and test splits. For those without a validation split, we can randomly partition the training data into training and validation splits. We note that in our paper we assume that public datasets have ``clean'' labels. We acknowledge that many widely used public datasets such as CIFAR10 or ImageNet~\citep{deng2009imagenet} may have mislabeled data points~\citep{northcutt2021pervasive}; however, the amount of label noise in these public datasets is significantly smaller than what the noisy label research community usually consider~\citep{han2018co,lee2019robust}, including our work. Therefore, we believe it is reasonable to consider the labels in public datasets as clean, i.e., less noisy, labels, and we do not expect the label noise in public datasets changes the conclusions in our paper.

\begin{figure}[ht]
  \begin{center}
    \includegraphics[width=0.65\textwidth]{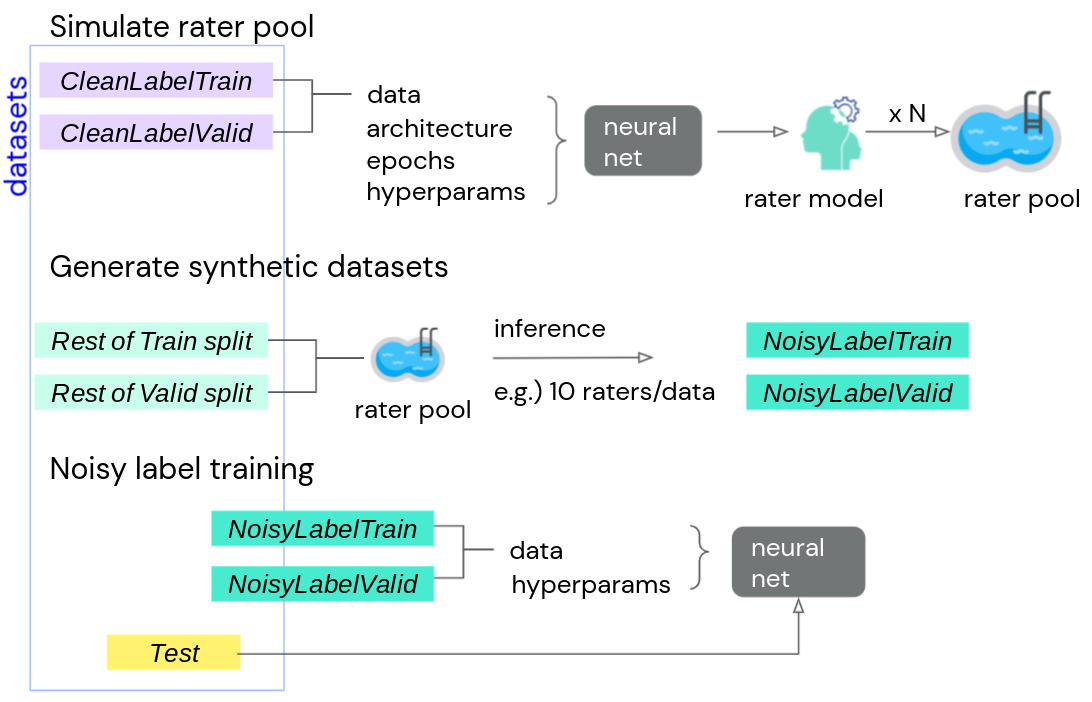}
  \end{center}
  \caption{Pseudo-labeling paradigm for simulating instance-dependent noisy labels.}\label{fig:framework_analogy}
\end{figure}

We further split the training and validation splits into two disjoint sets, respectively. More specifically, we partition the training set into \RaterTrain~and \SynthTrain, and the validation set into \RaterValid~and \SynthValid. We use the data in \RaterTrain~with clean labels to train a set of rater models, which can be any standard models for the problem domain. The data in the \RaterValid~split can be used to evaluate the rater models. For example, the test accuracy with respect to the clean labels on the \RaterValid~split can be used as a feature of a rater model. We can obtain a pool of rater models by choosing different architectures, training epochs, and other training configurations, which can all be used as \emph{rater features}.
Then we use all or a subset of models from the rater pool to run inference on the data in the \SynthTrain~and \SynthValid~splits. In this way we obtain multiple noisy labels for every data in these two splits, and we replace the clean labels with these noisy labels.
We note that in this paper, when we run inference using a rater model, we use the ``hard predictions'', i.e., each example is labeled according to the largest logit of the rater model's prediction. It is also valid to treat the output of the rater models as a distribution over the classes and sample a noisy label from it.
We find that in order to control the amount of label noise in these two splits, it is important to train a diverse set of rater models using different combinations of architectures, training steps, learning rate, and batch size. The details for the rater models that we use throughout this paper are provided in Appendix~\ref{apx:dataset_details}.
To perform label noise research, we can use the \SynthTrain~split to train models and use the \SynthValid~split for hyperparameter tuning.\footnote{The \SynthValid~split also contains noisy labels, which may affect the hyperparameters that we select. Understanding the impact of label noise in the validation set is beyond the scope of this paper and will be a future direction.}
For the \Test~split, we use the original clean labels. We illustrate our framework in Figure \ref{fig:framework_analogy}. To summarize, we split the dataset into 5 disjoint sets: 
\begin{itemize}\setlength\itemsep{0.3em}
    \item \RaterTrain: a set of data with clean labels, used for training rater models.
    \item \RaterValid: a set of data with clean labels, used for evaluating rater models.
    \item \SynthTrain: a set of data with multiple noisy labels (prediction of rater models), used for model training with noisy labels.
    \item \SynthValid: a set of data with multiple noisy labels (prediction of rater models), used for hyperparameter tuning when training on the \SynthTrain~split.
    \item \Test: a set of data with clean labels for final evaluation of the model trained on \SynthTrain.
\end{itemize}

In most of our experiments, the sizes of the \RaterTrain~and \SynthTrain~splits are around 50\% of the original training and validation splits. However, this ratio can be adjusted depending on the problem of interest. For a synthetic dataset with multiple noisy labels, i.e., the \SynthTrain~and \SynthValid~splits, we use the following two metrics to measure the amount of noise in the dataset: (1) overall rater error rate, which is defined as the fraction of the incorrect labels among all the labels given by all the raters, and (2) Krippendorff's alpha (\kalpha) \citep{kalpharef}, which measures the agreement between the raters. We note that the computation of Krippendorff's alpha does not require the clean labels. Usually, datasets with higher \kalpha~are less noisy. All model training in this and the following sections are performed on TPUs in our internal cluster.

\subsection{Dataset evaluation}\label{sec:dataset_eval}

Once we have the synthetic datasets, the next step is to compare them with other simulation methods. More specifically, we make comparison with independent random flipping and class-conditional random flipping, and show that the distribution of the noisy labels that we generate is closer to real human labels. We use the following metric named \emph{mean total variation distance} to measure the difference between the distribution of noisy labels in different datasets.

Let $D_1 = \{(x_i, y_i^1)\}_{i=1}^n$ and $D_2 = \{(x_i, y_i^2)\}_{i=1}^n$ be two noisy label datasets with the same set of input features. We consider soft labels, i.e., $y_i^1,y_i^2\in\mathbb{R}^K$ are probability distributions over $\{1,\ldots, K\}$. The mean total variation distance between datasets $D_1$ and $D_2$ is defined as $d_{TV}(D_1, D_2) := \frac{1}{2n}\sum_{i=1}^n \|y_i^1 - y_i^2\|_1$.

We use the CIFAR10-H dataset~\citep{peterson2019human} as the real human labels. This dataset contains the 10K data points from the CIFAR10 test split with around $50$ labels for each data. To create the synthetic noisy label datasets using our framework in Section~\ref{sec:dataset_gen}, we train rater models using the~\RaterTrain~ split and run inference on the CIFAR10 test data.\footnote{Notice that this is slightly different from Section~\ref{sec:dataset_gen} since the noisy labels are not generated on the \SynthTrain~or \SynthValid~splits, but on the test split. However, this is our only choice since CIFAR10-H only has human labels for the test split.} We create three synthetic datasets with low, medium and high amount of noise. We train $10$ rater models for each dataset and thus each example has $10$ noisy labels. The rater error rates---defined as the ratio of the number of incorrect labels to the total number of labels in the dataset---of the three datasets are $13.0\%$, $21.7\%$, and $ 52.4\%$, respectively. In the following, we call these three rater error rates the \emph{targeted error rates}. Note that when we compute the rater error rates, we corrected the mislabels data in the original CIFAR10 test split according to~\citet{northcutt2021pervasive}. Details for the rater models are presented in Appendix~\ref{apx:dataset_details}. With the three synthetic datasets, we then create other datasets for comparison. For \textbf{real human labels}, we notice that the CIFAR10-H dataset has a rater error rate around $4.8\%$, much lower than the amount of noise in our synthetic datasets. Since the goal of our framework is to create controllable noise level, we do not enforce the rate error rate of our datasets to match CIFAR10-H; instead, we upsample the incorrect labels in CIFAR10-H to match the rater error rates of our three datasets. The method of upsampling incorrect labels to create datasets with controllable amount of noise has been studied in~\citet{northcutt2021confident,northcutt2021pervasive}. Thus, we create noise-controlled human label datasets with the three targeted error rates. For \textbf{independent random flipping}, we generate three datasets by choosing $\delta$ to be each of the three targeted error rates and sampling $10$ noisy labels for each data. For \textbf{class-conditional random flipping}, for each synthetic dataset, we first compute its class confusion matrix, and then use it as the probability transition matrix $T$ for the class-conditional setup and then sample $10$ noisy labels for each data. Then, for each targeted error rate, we compute the mean total variation distance between the real human labels and datasets with independent random flipping, class-conditional random flipping, and our synthetic dataset, respectively. The results are provided in Table~\ref{tab:dataset_eval}. As we can see, for every noise level, the mean total variation distance between our synthetic dataset and the noise-controlled human labels is smaller than that of independent and class-conditional random flipping. Thus, we conclude that the distribution of noisy labels that our framework generates is closer to human labels compared to independent and class-conditional random flipping. 

\begin{table}[t]
\caption{Dataset evaluation. The numbers in the table are the $95\%$ confidence intervals of the mean total variation distance between the synthetic datasets generated by different methods and the noise-controlled human labeled datasets. Our datasets are closer to the human labels compared to independent and class-conditional random flipping.}\label{tab:dataset_eval}
\begin{center}
\setlength\tabcolsep{9pt}
\begin{tabular}{ c|c|c|c } 
 \toprule
 noise level & low & medium & high  \\
 \midrule
 independent flipping & $0.207 \pm 0.005$ &
$0.335 \pm 0.008$ &
$0.787 \pm 0.019$
 \\ 
 class-conditional & $0.196 \pm 0.005$ &
$0.320 \pm 0.008$ &
$0.766 \pm 0.019$
 \\ 
 ours & \textbf{0.180} $\pm$ \textbf{0.005} &
\textbf{0.301} $\pm$ \textbf{0.008} &
\textbf{0.742} $\pm$ \textbf{0.019}
 \\
 \bottomrule
\end{tabular}
\end{center}
\end{table}

\section{Impact of Label Noise on Deep Learning Models}\label{sec:impact_analysis}

With the instance-dependent synthetic datasets with noisy labels, our next step is to study the impact of noisy labels on deep learning models. Interestingly, there exist different views for the impact of noisy labels to deep neural networks. While most of the recent research works on noisy labels try to design algorithms that can tackle the negative impact of label noise, some other works claim that deep learning models are robust to independent random label noise \citep{rolnick2017deep,li2020gradient} without using sophisticated algorithms. A prominent example is the weak supervision paradigm \citep{ratner2016data,ratner2017snorkel}, where massive training datasets are generated by weak raters and labeling functions. Other lines research indicate that large neural network can easily fit all the noisy labels in the training data~\citep{zhang2021understanding}, while smaller models may be more robust against label noise due to the regularization effect~\citep{advani2020high,belkin2019reconciling,northcutt2021pervasive}.

We hypothesize that the negative impact of noisy labels is problem-dependent. While in most cases the incorrect labels can impair models’ performance, the impact may depend on factors related to the data distribution and the model. In this section, we choose the following factors to measure the impact of label noise: the class imbalance, the inductive bias of the model (in particular, pretraining vs random initialization), and the difficulty of the task (test accuracy that models can achieve when clean labels are accessible). Note that for better understanding, we decouple these factors with algorithm design: In this section, we choose simple SGD-style training algorithms with cross-entropy loss and focus on analyzing the impact of label noise; the discussion on more sophisticated algorithms to tackle label noise is presented in Sections~\ref{sec:benchmark} and~\ref{sec:lqm}.
We do not aim to study label aggregation methods either. Instead, in this and the following sections, given a synthetic dataset with multiple noisy labels, we generate a dataset with a \emph{single noisy label} by independently and uniformly selecting a random noisy label for every data point. This is a simulation of the practical setting where we have a pool of raters and for each data, we choose a random rater from the pool and request a label.

\subsection{Label Noise Has Higher Impact on More Imbalanced Datasets}\label{sec:imbalance}

\begin{figure*}[ht]
\centering
{\includegraphics[width=0.24\textwidth]{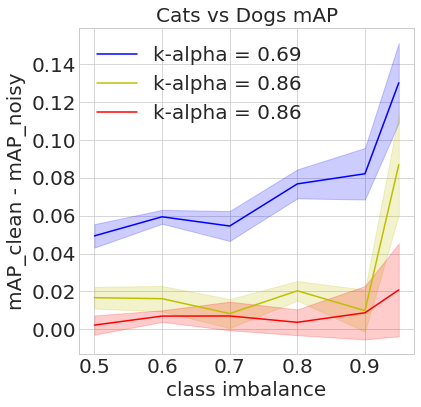}}
{\includegraphics[width=0.24\textwidth]{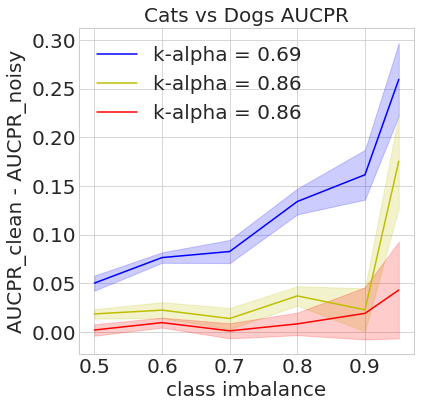}} 
{\includegraphics[width=0.24\textwidth]{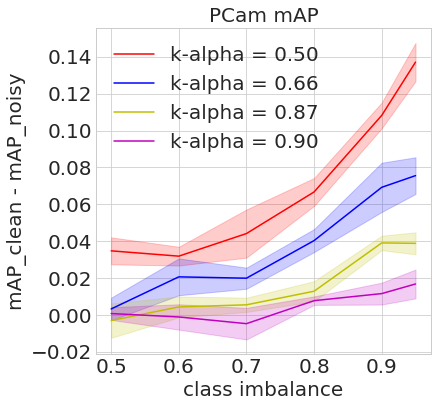}}
{\includegraphics[width=0.24\textwidth]{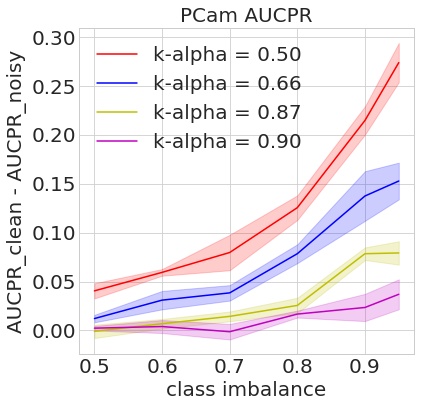}}
\caption{Impact of label noise for tasks with different class imbalance. The x-axis represents class imbalance, measured by the fraction of the majority class. For PCam, we use MobileNet-v1 \citep{howard2017mobilenets} model, and for Cats vs Dogs, we use ResNet50 \citep{he2016deep}. The \kalpha s correspond to the synthetic datasets before subsampling.}
\label{fig:imbalance_small}
\end{figure*}

One of the important characteristics of many real world datasets is that the classes are usually imbalanced. When the classes are more imbalanced, the impact of noisy labels may become more pronounced since the number of data in the minority classes is already small, and noisy labels can further corrupt these data, making the learning procedure more difficult. We validate this hypothesis in this section.
We use two binary classification tasks, PatchCamelyon (PCam) \citep{veeling2018rotation,bejnordi2017diagnostic} and Cats vs Dogs (CvD) \citep{elson2007asirra}. We generate synthetic noisy label datasets with different \kalpha s, and for each of these datasets, we subsample the two classes to create several smaller datasets with different class imbalance but the same total number of data. We note that here we control the class imbalance to be the same for all of the \SynthTrain, \SynthValid, and \Test~splits. We train models with clean and noisy labels and use the difference in mean average precision (mAP)~\citep{Zhang2009} and area under the precision-recall curve (AUCPR)~\citep{raghavan1989critical} as the indicators for the impact of label noise. The results are shown in Figure~\ref{fig:imbalance_small}. As we can see, the impact of label noise becomes more significant as the classes become more imbalanced.

\subsection{Pretraining Improves Robustness to Label Noise}\label{sec:pretrain}
One model training technique that is often used in practice, especially for computer vision and natural language tasks, is to pretrain the models on some large benchmark datasets and then fine-tune them using the data for specific tasks. It has been observed that model pretraining can improve robustness to independent random label noise \citep{hendrycks2019using} and the web label noise considered by \citet{jiang2020beyond}. Here we show that this can still be observed in our synthetic framework. A simple explanation is that model pretraining adds strong inductive bias to the models and thus they are less sensitive to a fraction of noisy labels during fine-tuning.

\begin{figure*}[ht]
\centering
{\includegraphics[width=0.48\textwidth]{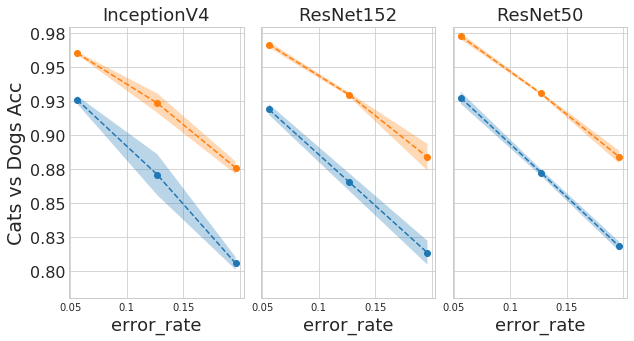}}
{\includegraphics[width=0.48\textwidth]{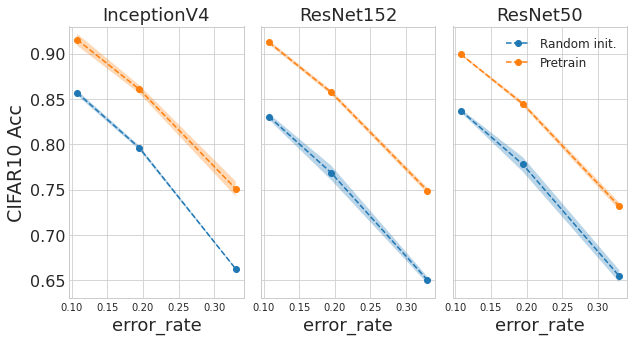}}
\caption{Pretrained models achieve better test accuracy on CvD and CIFAR10. Moreover, as the amount of label noise increases, the amount of test accuracy drop is smaller for pretraining (e.g. the slope is smaller) than training from random initialization.}
\label{fig:pretraining}
\end{figure*}

We validate this hypothesis using two datasets, Cats vs Dogs (CvD) and CIFAR10. For both datasets, we generate three synthetic noisy label datasets using our framework with different rater error rates. We compare the test accuracy on the \Test~split (with clean labels) between the models that are trained from random initialization and those that are fine-tuned from models pretrained on ImageNet \citep{deng2009imagenet}. We experiment with three different architectures, including Inception-v4 \citep{szegedy2017inception}, ResNet152, and ResNet50 \citep{he2016deep}. As we can see in Figure~\ref{fig:pretraining}, models that are pretrained on ImageNet achieve better test accuracy. In addition, for pretrained models, the test accuracy tends to drop more slowly compared to models that are trained from random initialization as we increase the amount of noise (rater error rate).

Meanwhile, we also observe that ImageNet pretraining does not improve the test accuracy under noisy labels for the PatchCamelyon dataset. This can be explained by the fact that the PatchCamelyon dataset consists of histopathologic scans of lymph node sections, and these medical images have very different distribution from the data in ImageNet. Therefore, the inductive bias that the model learned from ImageNet pretraining may not be helpful on PCam.

\subsection{Easier Tasks Are More Sensitive to Label Noise}\label{sec:task_difficulty}
We also study the impact of label noise on tasks with different difficulty levels (the test accuracy models can achieve when clean labels are accessible). Our hypothesis here is that when a task is already hard to learn even given clean labels, then the impact of label noise is smaller. The reason can be that when a classification task is hard, the data distributions of different classes are relatively close such that even if some data are mislabeled, the final performance may not be heavily impacted. On the contrary, label noise may be more detrimental to easier tasks as the data distribution can significantly change when well-separated data points get mislabeled. We validate this hypothesis with two experiments.

\begin{figure*}[ht]
\centering
{\includegraphics[width=0.4\textwidth]{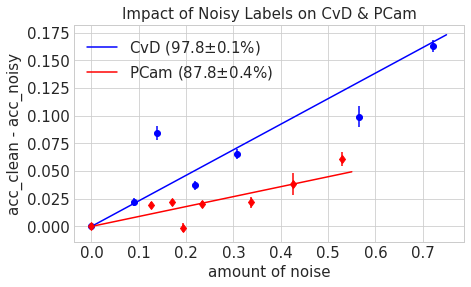}}
{\includegraphics[width=0.4\textwidth]{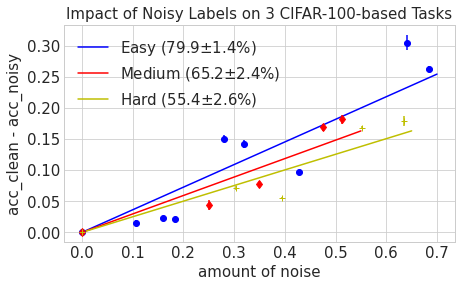}}
\caption{Impact of label noise on tasks with different difficulty levels. The numbers in the legend correspond to test accuracies when training with clean labels for every task. The x-axis represents the amount of noise, measured by $1.0-$\kalpha. The y-axis represents the negative impact of label noise, measured by the difference in test accuracy when training with clean and noisy labels. Scattered points represent the pairs of noise level and the accuracy drop, and solid lines show the linear fit of the scattered points in the same color.}
\label{fig:difficulty}
\end{figure*}

\paragraph*{Setup}
Our first experiment involves two binary classification tasks, i.e., PatchCamelyon (PCam) with MobileNet-v1 \citep{howard2017mobilenets} and Cats vs Dogs with ResNet50~\citep{he2016deep}. We generate synthetic noisy label datasets with different \kalpha s using our framework, and compare the accuracies when the models are trained with clean and noisy labels. We observe that CvD is easier than PCam (clean label accuracy 97.8 $\pm$ 0.1\% vs 87.7 $\pm$ 0.4\%). In our second experiment, we design three 20-way classification tasks with the same number of data but different difficulty levels by subsampling different classes from the CIFAR100 dataset. We call the three tasks the \emph{easy}, \emph{medium}, and \emph{hard} tasks. Details for the design of the three tasks are provided in Appendix~\ref{apx:difficulty}, and we observe that with clean labels, we can obtain test accuracies of 79.9 $\pm$ 1.4\%, 65.2 $\pm$ 2.4\%, and 55.4 $\pm$ 2.6\% for the three tasks, respectively. We generate synthetic datasets with different amounts of noise, measured by \kalpha~and use the MobileNet-v2 model \citep{sandler2018mobilenetv2}.

\paragraph*{Results}
We study the impact of label by measuring the absolute difference in test accuracy when training with clean and noisy labels. The results are shown in Figure~\ref{fig:difficulty}. As we can see, the impact of noisy labels is higher on the easy task: On CvD, the drop in test accuracy grows faster as we increase the amount of label noise (indicated by $1.0-$\kalpha) compared to PCam, and similar phenomenon can be observed on the three CIFAR100-based tasks.

\section{Benchmarking Noisy Label Algorithms}\label{sec:benchmark}

With our instance-dependent synthetic noisy label datasets, a natural follow-up question is how existing techniques for mitigating the impact of label noise perform on our benchmarks. In particular, we are interested in the difference of the algorithms' performance when using our synthetic datasets and using noisy label datasets with independent random label noise. In this and the next section, when we mention a dataset uses random label noise, we mean with certain probability (rater error rate), the label of each data point is flipped to an incorrect label that is uniformly selected. This flipping event is independent of other data points and the image itself.

\subsection{Experiment Setup}\label{sec:benchmark_experiment_setup}
We compare the following $5$ algorithms: vanilla training with cross-entropy loss (Baseline), Bootstrap~\citep{reed2014training}, Co-Teaching~\citep{han2018co}, cross-entropy loss with Monte Carlo sampling (MCSoftMax)~\citep{collier2020simple}, and MentorMix~\citep{jiang2020beyond}\footnote{We also experimented with F-Correction~\citep{patrini2017making} and RoG~\citep{lee2019robust} but did not observe significant improvement over the baseline on our synthetic datasets. Thus, we choose to not report the results of these two algorithms.} on $4$ tasks: CIFAR10, CIFAR100, PatchCamelyon, and Cats vs Dogs.\footnote{We also generated synthetic datasets using ImageNet. However, none of the noisy label techniques performs significantly better than vanilla training with cross entropy loss, thus we do not present the results here.} For each task, we generate $3$ synthetic noisy label datasets with different amount of noise using our framework. According to the rater error rate, the noisy label datasets are marked as ``low'', ``medium'', and ``high'' in Figures~\ref{fig:benchmark_acc} and~\ref{fig:benchmark_diff}. Details for these datasets can be found in Appendix~\ref{apx:dataset_details}. For each of our synthetic dataset, we generate another dataset that uses random label noise and has the \emph{same} rater error rate. We compare the performance of the $5$ algorithms on these paired datasets, and aim to measure the difficulty of noisy label datasets when the label errors are generated using our framework or independent random flipping. All the experiments use the ResNet50 architecture.

\subsection{Results}\label{sec:benchmark_results}
Interestingly, we find different behavior for tasks with different number of classes. For tasks with a large number of classes such as CIFAR100, we find that most algorithms achieve better test accuracy on our synthetic datasets compared to random label noise. On binary classification problems such as PatchCamelyon and Cats vs Dogs, however, the trend is opposite, i.e., most algorithms perform worse on our synthetic datasets. On CIFAR10, we observe mixed behavior: depending on the amount of noise and the algorithm, the test accuracy can be higher either on our synthetic datasets or those with random label noise. The results are shown in Figure~\ref{fig:benchmark_acc}, and exact numbers are provided in Appendix~\ref{apx:benchmark_results}.

\begin{figure*}[ht]
\centering
{\includegraphics[width=0.95\textwidth]{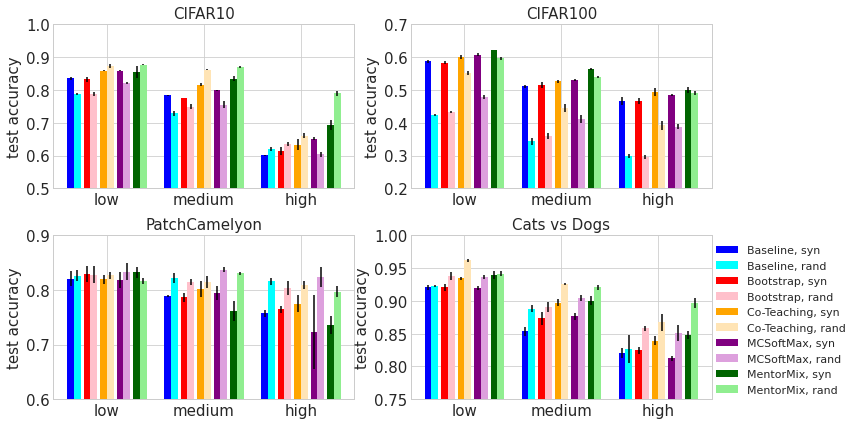}}
\caption{Benchmarking noisy label algorithms using our synthetic dataset and random label noise. Each pair of adjacent bars shows test accuracy on two datasets: our synthetic dataset (darker color) and random label noise (lighter color) with the same rater error rate. On CIFAR100, our datasets are easier than random noise, while on binary classification tasks (PCam and CvD), our datasets are harder.}
\label{fig:benchmark_acc}
\end{figure*}

\begin{figure*}[ht]
\centering
{\includegraphics[width=0.9\textwidth]{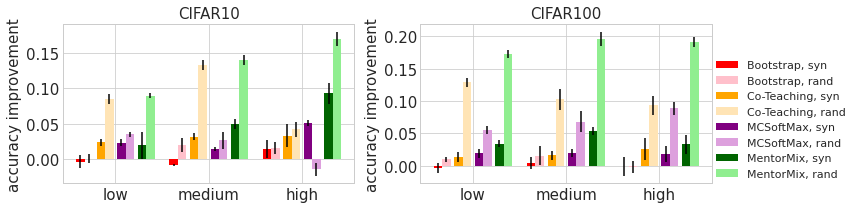}}
\caption{Improvement in test accuracy using noisy label techniques. Each pair of adjacent bars shows test accuracy improvement compared to the baseline on two datasets: our synthetic dataset (darker color) and random label noise (lighter color) with the same rater error rate. In most cases, the accuracy improvement tends to be smaller under our synthetic framework.}
\label{fig:benchmark_diff}
\end{figure*}

This phenomenon can be explained as follows.
For binary classification problems, in our synthetic framework, the mislabeled data are usually the ambiguous ones that located around the decision boundary. This label noise can hurt the models' performance more since the important information around the decision boundary is corrupted.
On the contrary, for tasks with a large number of classes, especially those with tree-structured classes involving a relatively small number of high level super classes and low level fine-grained classes, such as CIFAR100, in our instance-dependent simulation framework, the label mistakes are usually among similar classes. For example, an image of a certain type of mammal may be mislabeled as another mammal, but it is unlikely to be labeled as a type of vehicles. In other words, the corruption of decision boundary only happens to similar fine-grained classes in our framework. Thus, given the same fractions of incorrect labels are the same, our synthetic label noise hurts the models' performance less compared to random noise. 

Another observation is that on CIFAR10 and CIFAR100, the performance improvement obtained by noisy label algorithms when compared with the baseline is usually smaller with our synthetic datasets. The performance improvement is presented in Figure~\ref{fig:benchmark_diff}.

We emphasize that our results demonstrate the importance of using instance-dependent synthetic benchmarks in the research on label noise: existing algorithms exhibit different behavior on our synthetic framework and random label noise, even if the fraction of mislabeled data is kept the same, and the performance gain observed using random label noise may not directly translate to the setting that we tested.

\section{Leveraging Rater Features: Label Quality Model}\label{sec:lqm}

Existing work in the noisy label literature commonly assumes that training labels are the only output of the data curation process. In practice however, the data curation process often produces a myriad of additional features that can be leveraged in downstream training, e.g., which rater is responsible for a given label, as well as that rater’s tenure, historical errors, and time spent on a given task. With our proposed method of simulating instance-dependent noisy labels via rater models, we can additionally simulate these rater features by extracting metadata from the rater models, e.g., the number of epochs used to train the rater models is a proxy for rater tenure. Another common practice in label curation is assigning multiple raters for a single example. This is commonly used to reduce the label noise via aggregation, or to evaluate the performance of individual raters against the pool. This practice assumes that agreement among multiple raters are more accurate than individual responses. 

With understanding of practical data collection setup, we introduce a technique for training with noisy labels, which we coin \emph{Label Quality Model} (LQM). LQM is an intermediate supervised task aimed at predicting the clean labels from noisy labels by leveraging rater features and a paired subset for supervision. The LQM technique assumes the existence of rater features and a subset of training data with both noisy and clean labels, which we call \emph{paired-subset}. We expect that in real world scenarios some level of label noise may be unavoidable. LQM approach still works as long as the clean(er) label is less noisy than a label from a rater that is randomly selected from the pool, e.g., clean labels can be from either expert raters or aggregation of multiple raters. LQM is trained on the paired-subset using rater features and noisy label as input, and inferred on the entire training corpus. The output of LQM is used during model training as a more accurate alternative to the noisy labels.

The intuition for LQM is to correct the labels in a rater-dependent manner. This means that by learning the patterns from the paired-subset, we can conduct rater-dependent label correction. For example, LQM can potentially learn that raters with a certain feature often mislabel two breeds of dogs, then it can possibly correct these two labels from similar raters for the rest of the data. Below we formally present the details of LQM. 

\subsection{Algorithm Design}\label{sec:lqm_algo}

Formally, let $D:= \{(x_i, y_i, r_i)\}_{i=1}^N$ be a noisy label dataset, e.g., the \SynthTrain~split,\footnote{As mentioned in Section~\ref{sec:dataset_gen}, the size of the \SynthTrain~split is around 50\% of the training split of the original dataset.} where $x_i$ is the input, $y_i$ is the one-hot encoded noisy label, and $r_i$ is the rater feature corresponding to $y_i$. Let $D_{ps} = \{(x_j, y_j, r_j, y^*_j)\}_{j=1}^M$ be the paired-subset, and $y^*_j$ be a more accurate label than $y_j$. We usually have $M \ll N$. We propose to optimize a parameterized model $\LQM(\theta; x, r, y)$ to approximate the conditional probability $P(y^* | x, r, y)$ using $D_{ps}$. We note that LQM leverages all the information from the input $x$, rater features $r$, and the noisy label $y$. 

Once we have the LQM, we proceed to tackle the main task using the noisy label dataset $D$. Instead of trying to predict $P(y_i | x_i)$, we replace the noisy labels $y_i$ with the outputs of LQM and train a model to predict $P(\LQM(\theta; x_i, r_i, y_i) | x_i)$. From experimentation, we find that by interpolating between noisy label $y_i$ and the output of LQM produces even stronger results. Therefore, we recommend training with target $\tilde{y_i} = \gamma \LQM(\theta; x_i, r_i, y_i) + (1-\gamma)y_i$, where $\gamma$ is a hyperparameter between $0$ and $1$ and can be selected using the validation set.
This is particularly helpful for datasets with a large number of classes such as CIFAR100, since it prevents the training target from getting too far from the original labels $y_i$.
Moreover, since $\tilde{y_i}$ specifies a distribution over the labels, we can also sample a single one-hot label according to the distribution $\tilde{y_i}$ as the target.

We use a small set of rater features in the simulated framework, such as the accuracy of the rater model on \RaterValid, the number of epochs trained, and the type of architecture.
In addition, we also use the paired-subset to empirically calculate the confusion matrix for each rater and use it as a feature for the rater.
Instead of training LQM with raw input $x$, we first train an auxiliary image classifier $f(x)$ and train LQM using the output logits of $f(x)$. The auxiliary classifier can be trained over either the full noisy dataset $D$ or the paired-subset $D_{ps}$. We find that the better option depends on the task and the amount of noise present. In our experimentation, we train $f(x)$ on both dataset options and select the better one. Given that LQM has fewer training examples, using an auxiliary image classifier significantly simplifies training.

\subsection{Experiment Setup and Results}\label{sec:lqm_setup}
For uniformity, we assume $M = 0.1N$, i.e. 10\% of training data has access to a clean label in all of our following experiments. For the main prediction model, i.e., $P(\LQM(\theta; x_i, r_i, y_i) | x_i)$, we use the ResNet50 architecture. For the auxiliary model $f(x)$, we use MobileNet-v2. The LQM itself is trained using a one-hidden-layer MLP architecture with cross-entropy loss. The number of hidden units in the MLP is chosen in $\{8, 16, 32\}$ as a hyperparameter. We conduct the following two experiments, and the exact numbers for the results are provided in Appendix~\ref{apx:benchmark_results}.

\begin{figure*}[ht]
\centering
{\includegraphics[width=0.78\textwidth]{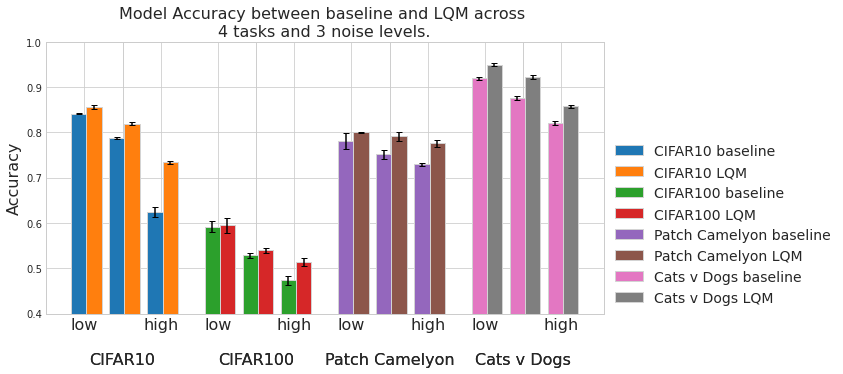}}
\caption{Training with LQM adjusted labels outperforms baselines across all datasets. The improvement from LQM is more prominent in medium and high noise settings. LQM models are trained by stochastically sampling a one-hot label from LQM output distribution, and tuning the $\gamma$ hyperparameter.}
\label{fig:lqm}
\end{figure*}

\paragraph*{LQM vs Baseline}
First, we compare the performance of models trained with LQM and the baseline models that are trained using vanilla cross-entropy loss without leveraging rater features. Since LQM has access to clean labels of 10\% of the data, for fair comparison, we ensure that the baseline models also have access to the same number of clean labels. The comparison is presented in Figure~\ref{fig:lqm}. As we can see, with rater features and the label correction step, in many cases, especially in the medium and high noise settings, LQM outperforms the baseline.

\begin{figure*}[ht]
\centering
{\includegraphics[width=0.95\textwidth]{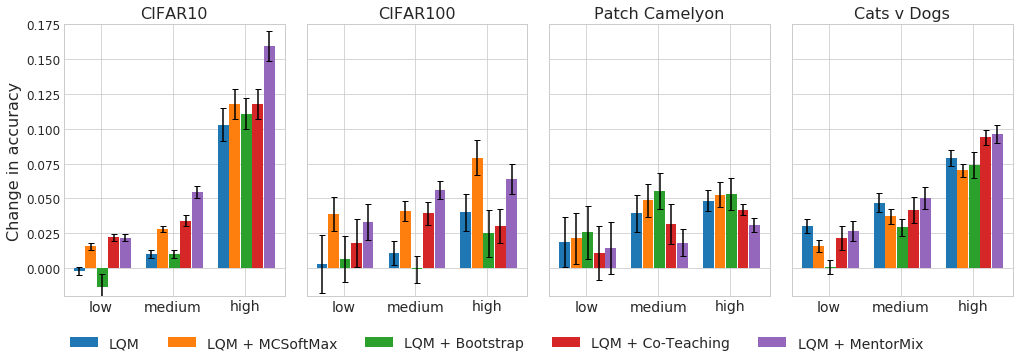}}
\caption{Accuracy improvement of LQM and the combinations of LQM and other techniques compared to the baseline. In many settings (CIFAR10, CIFAR100 and high noise setting in CvD), combining other techniques with LQM further improves the test accuracy. In other cases (PCam and low/medium noise for CvD), the performance gain is less significant.}
\label{fig:lqm_plus}
\end{figure*}

\paragraph*{Combining LQM with Other Techniques}
In the second experiment, we investigate the conjunction of LQM with other noisy labels techniques. We hypothesize that, depending on the technique, LQM may be correcting a different kind of noise from existing techniques, and thus can potentially lead to further performance improvement. To combine LQM with another technique, we sample a hard label from the soft distribution specified by $\tilde{y_i}$, and apply other noisy labels techniques on top of the sampled hard label. We consider the same set of noisy labels techniques as the previous section. We find that on CIFAR10 and CIFAR100, combining LQM with other techniques usually lead to further performance improvement. The improvement can also be observed in the high noise setting for CvD. The results are illustrated in Figure~\ref{fig:lqm_plus}.

As a final note, since LQM assumes access to a subset of data with clean labels, and also uses an auxiliary classifier $f(x)$, it has some similarity with semi-supervised learning (SSL). We notice that several state-of-the-art SSL techniques such as FixMatch~\citep{sohn2020fixmatch}, UDA~\cite{xie2019unsupervised}, self-training with noisy student~\citep{xie2020self} use specifically designed data augmentations that are only suitable for specific types of data, whereas LQM can be applied to any type of data as long as we have rater features. We also expect that combining certain SSL techniques (e.g., data augmentation and consistency training) can further improve the results; however, these extensions are beyond the scope of this paper.

\section{Additional Related Work}\label{sec:related_work}
There is a large body of literature on learning with noisy labels. We mentioned several related work in the previous sections and it is certainly not an exhaustive list. Since we focus on simulation frameworks for noisy label research, we first review prior works that try to simulate noisy labels using methods beyond random label flipping or permutation. As mentioned in previous sections, we are aware that several prior works~\citep{lee2019robust,robinson2020strength,berthon2021confidence,zhu2021second,zhang2021learning} also use similar pseudo-labeling paradigm to generate synthetic datasets with noisy labels. \citet{seo2019combinatorial} use a similar idea of nearest neighbor search in the feature space of a pretrained model with clean labels to generate noisy labels. Compared with these works, our study is much more comprehensive with a diverse set of tasks and model architectures. We conduct a series of analysis on the impact of noisy labels, and propose a method to generate synthetic rater features and use them for improving robustness. These points were not considered in the two prior works. Other approaches to simulating label noise have also been studied in the literature. \citet{jiang2020beyond} proposes a framework to generate controlled \emph{web label noise}, in which case new images with noisy labels are crawled from the web and then inserted to an existing dataset with clean labels. The framework differs from our approach and the two frameworks should be considered complementary for generating noisy label datasets. In particular, the method by \citet{jiang2020beyond} is more suitable for web-based data collection, e.g., WebVision~\citep{li2017webvision} whereas ours is more suitable for simulating human raters. Moreover, \citet{wang2018iterative} and \citet{seo2019combinatorial} consider open-set noisy labels, where the mislabeled data may not belong to any class of the dataset, similar to \citet{jiang2020beyond}. Another approach to generating datasets with controllable about of label noise is to first identify a dataset with noisy labels (potentially some public datasets~\citep{northcutt2021confident,northcutt2021pervasive}) and then use the confident learning (CL) method~\citep{northcutt2021confident} to increase or decrease the amount of label noise proportionally to the distribution of real-world label noise in the dataset. The idea is to model the joint distribution of noisy and true labels and then generate the noisy labels based on the noise-increased or noise-decreased joint distribution of noisy and true labels. This differs from our method since we use rater models, which are trained neural networks to generate noisy labels for each instance. 

Tackling noisy labels using a small subset of data with clean labels has been considered in a few prior works. Common approaches include pretraining or fine-tuning the network using clean labels \citep{xiao2015learning,krause2016unreasonable}, and distillation~\citep{li2017learning}. In loss correction approaches, a subset of clean labels are often used for estimating the confusion matrix~\citep{patrini2017making,hendrycks2018using}. \citet{veit2017learning} propose a method that estimates the residuals between the noisy and clean labels. \citet{ren2018learning} use the clean label dataset to learn to reweight the examples. \citet{tsai2019countering} combine clean data with self-supervision to learn robust representations. Our approach differs from these prior works since we leverage the additional rater features to learn an auxiliary model that corrects noisy labels in a rater-dependent manner, and can be combined with other techniques to further improve the performance as shown in Section~\ref{sec:lqm}. 
Learning from multiple annotators has been a longstanding research topic. The seminal work by~\citet{dawid1979maximum} uses the EM algorithm to estimate rater reliability, and much progress has been made since then~\citep{raykar2010learning,zhang2014spectral,lakshminarayanan2013inferring}. Rater features is commonly available in many human annotation process. In crowdsourcing, several prior work focus on estimating the reliability of raters~\citep{raykar2010learning,tarasov2014dynamic,moayedikia2019improving}, and rater aggregation~\citep{vargo2003learning,chen2013pairwise}. Item response theory~\citep{embretson2013item} from the psychometrics literature uses a latent-trait model to estimate the proficiency of raters and the difficulty of examples, and has a similar underlying principle to our work.

Our method is also broadly related to several other lines of research. Training a pool of rater models is similar to ensemble method~\citep{dietterich2000ensemble}, which is usually used to boost test accuracy~\citep{freund1997decision} or improve uncertainty estimation~\citep{lakshminarayanan2017simple}. Training new models using noisy labels provided by the rater models is similar to knowledge distillation~\citep{bucilua2006model,hinton2015distilling}. Designing instance-dependent noisy label generation framework can be considered as reducing underspecification~\citep{d2020underspecification} in generating label noise.

\section{Conclusions and Limitations}\label{sec:conclusion}
In this paper, we propose framework for simulating instance-dependent label noise. Our method is based on the pseudo-labeling paradigm. We show that the distribution of noisy labels in our synthetic datasets is closer to human labels compared to independent random label noise. With our synthetic datasets, we evaluate the negative impact of label noise on deep learning models, and demonstrate that label noise is more detrimental under class imbalanced settings, when pretraining is not used, and on easier tasks.
We observe that existing algorithms for learning with noisy labels exhibit different behavior on our synthetic datasets and the datasets with random label noise. Using the rater features from our simulation framework, we propose a new technique, Label Quality Model, that leverages annotator features to predict and correct against noisy labels. We show that our technique can be combined with existing approaches to further improve model performance.

Our work demonstrates the importance of using instance-dependent datasets for noisy label research. As noted above, the performance gain of noisy label techniques on a dataset with independent random label flipping may not directly transfer to our synthetic datasets. We hope our framework can serve as an option for the noisy label research community to develop more efficient methods for practical challenges.

Several limitations of our framework: 1) As discussed in Section~\ref{sec:related_work}, it focuses more on simulating human errors, whereas there might be other types of label noise in practical settings that need other simulation methods; 2) Controlling the amount of label noise in the datasets requires careful archtecture selection and hyperparameter tuning, thus is harder compared to random flipping methods; 3) LQM requires the paired-subset containing both clean and noisy labels. This requirement may not be satisfied in some applications.

\section*{Acknowledgements}
The authors would like to thank Dilan G\"or\"ur, Jonathan Uesato, Timothy Mann, and Denny Zhou for helpful discussions.

% Bibliography components
% abbrvnat
\bibliographystyle{abbrvnat}
\setlength{\bibsep}{5pt} 
\setlength{\bibhang}{0pt}
\bibliography{ref.bib}

\newpage
\appendix
\section*{Appendix}
\section{Details of Synthetic Datasets}\label{apx:dataset_details}
In this section, we provide more details of the synthetic data generation process. In particular, we provide the architectures and hyperparameters of the rater models in these datasets. All the models use standard cosine learning rate decay schedule, as well as the standard flipping and cropping data augmentation.
In the following, for rater models that use the same architecture, they are randomly initialized independently.

CIFAR10 datasets in Section~\ref{sec:dataset_eval}. We use $10$ rater models for the low, medium, and high noise datasets, respectively. \emph{Low noise}: $3$ Inception-v1~\citep{szegedy2015going}, $1$ Inception-v3~\citep{szegedy2016rethinking}, $2$ Inception-ResNet-v2~\citep{szegedy2017inception}, $2$ MobileNet-v1~\citep{howard2017mobilenets}, $2$ VGG16~\citep{simonyan2014very} models. The models are trained with batch size $256$, $80,000$ steps, and initial learning rate $0.01$. \emph{Medium noise}: $2$ Inception-v4, $2$ MobileNet-v1, $2$ MobileNet-v2, $1$ NASNetMobile~\citep{zoph2018learning}, $1$ ResNet50, $1$ ResNet101, $1$ VGG16. All models are trained with batch size $256$, initial learning rate $0.01$ and $17,000$ steps. \emph{High noise}: $2$ Inception-v2, $1$ Inception-ResNet-v2, $2$ MobileNet-v1, $1$ MobileNet-v2, $2$ ResNet50, $1$ ResNet101, $1$ ResNet152. All models are trained with batch size $256$, initial learning rate $0.01$ and $5,000$ steps.

For the PCam dataset in Section~\ref{sec:task_difficulty}, we use $20$ rater models involving $10$ architectures: Inception-v1, Inception-v2~\citep{szegedy2016rethinking}, Inception-v3, Inception-v4~\citep{szegedy2017inception}, MobileNet-v1, MobileNet-v2~\citep{sandler2018mobilenetv2}, ResNet50, ResNet152~\citep{he2016deep}, VGG16, VGG19. For each architecture, we use two different initial learning rates: $0.01$ and $0.001$ to train two different models. All the models are trained with batch size $256$ and $10,000$ steps.

For the CvD dataset in Section~\ref{sec:task_difficulty}, we use $10$ rater models, involving the same $10$ architectures in the PCam dataset mentioned above. All the models are trained with batch size $128$, initial learning rate $0.001$, and $10,000$ steps.

For the \emph{Easy} task based on CIFAR100 in Section~\ref{sec:task_difficulty}, we use $10$ rater models with the following architectures:
Inception-v1, Inception-v2, Inception-v3, Inception-v4, Inception-ResNet-v2, MobileNet-v1, MobileNet-v2, ResNet50, ResNet101, ResNet152. We use batch size $128$, initial learning rate $3\times 10^{-5}$, and $5,000$ training steps. For the \emph{Medium} task, we use $11$ rater models, each using its own architecture. The $11$ architectures include the $10$ architectures for the PCam dataset in Section~\ref{sec:task_difficulty} with an additional ResNet101. We use batch size $128$, initial learning rate $0.003$, and $40,000$ steps. The \emph{Hard} task uses $11$ rater models with the same architectures as the \emph{Medium} task, with batch size $128$, initial learning rate $0.01$ and $2\times 10^5$ steps.

For the $3$ CvD datasets in Section~\ref{sec:imbalance}, we use $10$ rater models with the same architectures as the PCam dataset in Section~\ref{sec:task_difficulty}. All models are trained with batch size $128$. For the three datasets, the (initial learning rate, number of steps) pairs are $(1\times 10^{-2}, 5\times 10^4)$, $(1\times 10^{-3}, 2.5\times 10^4)$, $(1\times 10^{-3}, 1\times 10^5)$, respectively.

For each of the $4$ PCam datasets in Section~\ref{sec:imbalance}, we use $20$ raters models, which uses the same combinations of the $10$ architectures and $2$ initial learning rates as in the PCam dataset in Section~\ref{sec:task_difficulty}. They all use batch size $256$. The $4$ datasets are generated by varying the number of training steps among $\{1, 2, 5, 8\}\times 10^4$.

In Section~\ref{sec:pretrain}, we use $3$ CvD datasets and $3$ CIFAR10 datasets. All the CvD datasets use $10$ rater models with the same set of architectures as the CvD dataset in Section~\ref{sec:task_difficulty}. All rater models are trained with batch size $128$. The (initial learning rate, number of steps) pairs are $(1\times 10^{-3}, 1\times 10^5)$, $(1\times 10^{-3}, 2.5\times 10^4)$, $(1\times 10^{-2}, 1\times 10^4)$, respectively. The CIFAR10 dataset with rater error rate $0.11$ is the same as the dataset in Section~\ref{sec:dataset_eval}. The CIFAR10 dataset with rater error rate $0.19$ uses the same rater models as the medium noise dataset in Section~\ref{sec:dataset_eval}. The CIFAR10 dataset with rater error rate $0.33$ uses the same rater models as the high noise dataset in Section~\ref{sec:dataset_eval} with the only difference being that the number of training steps is $12,000$.

In Section~\ref{sec:benchmark} and Section~\ref{sec:lqm}, we use $3$ datasets for each of the $4$ tasks. The rater error raters of these datasets are provided in the tables in Appendix~\ref{apx:benchmark_results}. Here, we provide details of the rater models in the synthetic datasets.

\textbf{CIFAR10} The same as Section~\ref{sec:dataset_eval}.

\textbf{CIFAR100} For all the $3$ CIFAR100 datasets, we use $11$ raters, with the same set of architectures as the \emph{Medium} and \emph{Hard} tasks in Section~\ref{sec:task_difficulty}. The (batch size, learning rate, number of steps) tuples for the low, medium, and high noise datasets are $(128, 1\times 10^{-3}, 1\times 10^4)$, $(256, 0.01, 2\times 10^5)$, and $(256, 0.01, 8\times 10^4)$, respectively.

\textbf{PCam} For all the $3$ PCam datasets, we use $20$ raters models (for the medium noise dataset, one of the Inception-v1 models failed due to system error, so we only have $19$ noisy labels for this dataset), which uses the same combinations of the $10$ architectures and $2$ initial learning rates as in the PCam dataset in Section~\ref{sec:task_difficulty}. They all use batch size $256$ and initial learning rate $0.01$. The number of steps are $3.5\times 10^4$, $1.5\times 10^4$, and $1\times 10^4$, for the low, medium, and high noise datasets, respectively.

\textbf{CvD} For all the $3$ CvD datasets, we use $10$ rater models with the same set of architectures as the CvD dataset in Section~\ref{sec:task_difficulty}. All rater models are trained with batch size $128$. The (initial learning rate, number of steps) pairs are $(1\times 10^{-3}, 5\times 10^4)$, $(1\times 10^{-2}, 1\times 10^4)$, $(1\times 10^{-3}, 1\times 10^4)$, for low, medium, and high noise, respectively.

\section{Details for Three CIFAR100-Based Datasets in Section~\ref{sec:task_difficulty}}\label{apx:difficulty}

As we know, the CIFAR100 dataset contains 20 super classes, each of which contains 5 fine-grained classes. We create the easy, medium, and hard tasks in the following way.

\begin{itemize}
    \item For the easy task, we select one fine-grained class from each of the 20 super classes, and form a 20-way classification task.
    
    \item For the medium task, we select 4 super classes that are semantically similar, i.e., large carnivores, large omnivores and herbivores, small mammals, and medium-sized mammals. We use all the fine-grained classes from these 4 super classes to form a 20-way classification task.
    
    \item For the hard task, we simply classify the 20 super classes, and we randomly subsample the data in order to match the total number of data points in the other two tasks. We note that this task is harder since the data in each super class is a mixture of 5 fine-grained classes. 
\end{itemize}

\section{Benchmarking Results}\label{apx:benchmark_results}

We provide exact numbers for the experimental results in Section~\ref{sec:benchmark} (Tables~\ref{tab:benchmark_cifar10},~\ref{tab:benchmark_cifar100},~\ref{tab:benchmark_pcam}, and~\ref{tab:benchmark_cvd}) and Section~\ref{sec:lqm} in the main paper (Tables~\ref{tab:lqm_plus_tech_cifar10},~\ref{tab:lqm_plus_tech_cifar100},~\ref{tab:lqm_plus_tech_pcam}, and~\ref{tab:lqm_plus_tech_cvd}).

\begin{table}[ht]
\caption{Test accuracy $\pm$ std (\%) of noisy label algorithms on CIFAR10}\label{tab:benchmark_cifar10}
\centering
\setlength\tabcolsep{9pt}
\resizebox{\linewidth}{!}{
\begin{tabular}{c|cc|cc|cc}
\toprule
noise	 & \multicolumn{2}{c|}{low (err=0.11)} & \multicolumn{2}{c|}{medium (err=0.19)} &
	 \multicolumn{2}{c}{high (err=0.48)} \\
dataset  &  synthetic & random &  synthetic & random &  synthetic & random	\\
\midrule
Baseline & 83.6 $\pm$ 0.5 & 78.7 $\pm$ 0.3 & 78.4 $\pm$ 0.1 & 72.9 $\pm$ 0.7 & 60.1 $\pm$ 0.2 & 61.9 $\pm$ 0.6 \\
Bootstrap & 83.2 $\pm$ 0.8 & 78.8 $\pm$ 0.6 & 77.6 $\pm$ 0.1 & 74.8 $\pm$ 0.7 & 61.5 $\pm$ 1.2 & 63.5 $\pm$ 0.7 \\
Co-Teaching & 85.9 $\pm$ 0.2 & 87.2 $\pm$ 0.6 & 81.6 $\pm$ 0.5 & 86.2 $\pm$ 0.2 & 63.4 $\pm$ 1.6 & 66.1 $\pm$ 0.9 \\
% F-Correction & 80.9 $\pm$ 5.0 & 80.2 $\pm$ 0.2 & 78.4 $\pm$ 0.4 & 79.4 $\pm$ 0.7 & 61.7 $\pm$ 2.8 & 66.2 $\pm$ 2.4 \\
MCSoftMax & 85.9 $\pm$ 0.1 & 82.2 $\pm$ 0.3 & 79.8 $\pm$ 0.2 & 75.5 $\pm$ 1.1 & 65.2 $\pm$ 0.4 & 60.4 $\pm$ 0.8 \\
MentorMix & 85.5 $\pm$ 1.9 & 87.7 $\pm$ 0.2 & 83.4 $\pm$ 0.7 & 86.9 $\pm$ 0.3 & 69.4 $\pm$ 1.5 & 78.9 $\pm$ 0.9 \\
\bottomrule
\end{tabular}
}
\end{table}

\begin{table}[ht]
\caption{Test accuracy $\pm$ std (\%) of noisy label algorithms on CIFAR100}\label{tab:benchmark_cifar100}
\centering
\setlength\tabcolsep{9pt}
\resizebox{\linewidth}{!}{
\begin{tabular}{c|cc|cc|cc}
\toprule
noise	 & \multicolumn{2}{c|}{low (err=0.25)} & \multicolumn{2}{c|}{medium (err=0.38)} &
	 \multicolumn{2}{c}{high (err=0.43)} \\
dataset  &  synthetic & random &  synthetic & random &  synthetic & random	\\
\midrule
Baseline & 58.8 $\pm$ 0.5 & 42.3 $\pm$ 0.3 & 51.1 $\pm$ 0.5 & 34.4 $\pm$ 1.1 & 46.7 $\pm$ 1.2 & 29.9 $\pm$ 0.6 \\
Bootstrap & 58.4 $\pm$ 0.6 & 43.3 $\pm$ 0.3 & 51.5 $\pm$ 0.8 & 35.9 $\pm$ 0.9 & 46.6 $\pm$ 0.9 & 29.6 $\pm$ 0.7 \\
Co-Teaching & 60.1 $\pm$ 0.6 & 55.2 $\pm$ 0.7 & 52.7 $\pm$ 0.5 & 44.6 $\pm$ 1.2 & 49.2 $\pm$ 1.2 & 39.2 $\pm$ 1.4 \\
MCSoftMax & 60.7 $\pm$ 0.4 & 47.8 $\pm$ 0.5 & 53.0 $\pm$ 0.4 & 41.2 $\pm$ 1.2 & 48.5 $\pm$ 0.3 & 38.8 $\pm$ 0.8 \\
MentorMix & 62.2 $\pm$ 0.1 & 59.6 $\pm$ 0.6 & 56.4 $\pm$ 0.3 & 53.9 $\pm$ 0.4 & 50.0 $\pm$ 0.8 & 49.0 $\pm$ 0.5 \\
\bottomrule
\end{tabular}
}
\end{table}

\begin{table}[ht]
\caption{Test accuracy $\pm$ std (\%) of noisy label algorithms on PatchCamelyon}\label{tab:benchmark_pcam}
\centering
\setlength\tabcolsep{9pt}
\resizebox{\linewidth}{!}{
\begin{tabular}{c|cc|cc|cc}
\toprule
noise	 & \multicolumn{2}{c|}{low (err=0.10)} & \multicolumn{2}{c|}{medium (err=0.18)} &
	 \multicolumn{2}{c}{high (err=0.23)} \\
dataset  &  synthetic & random &  synthetic & random &  synthetic & random	\\
\midrule
Baseline & 82.1 $\pm$ 1.3 & 82.6 $\pm$ 1.0 & 78.9 $\pm$ 0.2 & 82.2 $\pm$ 1.0 & 75.7 $\pm$ 0.7 & 81.6 $\pm$ 0.6 \\
Bootstrap & 82.9 $\pm$ 1.4 & 82.8 $\pm$ 1.6 & 78.6 $\pm$ 0.9 & 81.4 $\pm$ 0.5 & 76.4 $\pm$ 0.7 & 80.3 $\pm$ 1.3 \\
Co-Teaching & 82.0 $\pm$ 0.8 & 82.7 $\pm$ 0.6 & 80.1 $\pm$ 1.4 & 81.4 $\pm$ 1.2 & 77.5 $\pm$ 1.5 & 80.9 $\pm$ 0.8 \\
MCSoftMax & 81.9 $\pm$ 1.5 & 83.4 $\pm$ 1.6 & 79.4 $\pm$ 1.2 & 83.7 $\pm$ 0.5 & 72.3 $\pm$ 6.7 & 82.3 $\pm$ 1.8 \\
MentorMix & 83.2 $\pm$ 1.1 & 81.7 $\pm$ 0.6 & 76.2 $\pm$ 1.9 & 83.0 $\pm$ 0.3 & 73.6 $\pm$ 1.6 & 79.7 $\pm$ 1.0 \\
\bottomrule
\end{tabular}
}
\end{table}

\begin{table}[ht]
\caption{Test accuracy $\pm$ std (\%) of noisy label algorithms on Cats vs Dogs}\label{tab:benchmark_cvd}
\centering
\setlength\tabcolsep{9pt}
\resizebox{\linewidth}{!}{
\begin{tabular}{c|cc|cc|cc}
\toprule
noise	 & \multicolumn{2}{c|}{low (err=0.09)} & \multicolumn{2}{c|}{medium (err=0.20)} &
	 \multicolumn{2}{c}{high (err=0.29)} \\
dataset  &  synthetic & random &  synthetic & random &  synthetic & random	\\
\midrule
Baseline & 92.0 $\pm$ 0.4 & 92.3 $\pm$ 0.2 & 85.4 $\pm$ 0.7 & 88.8 $\pm$ 0.5 & 82.0 $\pm$ 0.7 & 82.7 $\pm$ 2.1 \\
Bootstrap & 92.1 $\pm$ 0.6 & 93.8 $\pm$ 0.6 & 87.3 $\pm$ 1.0 & 89.0 $\pm$ 0.8 & 82.4 $\pm$ 0.6 & 85.8 $\pm$ 0.4 \\
Co-Teaching & 93.4 $\pm$ 0.2 & 96.2 $\pm$ 0.2 & 89.7 $\pm$ 0.5 & 92.6 $\pm$ 0.2 & 84.0 $\pm$ 0.6 & 86.7 $\pm$ 1.3 \\
MCSoftMax & 92.0 $\pm$ 0.3 & 93.7 $\pm$ 0.3 & 87.6 $\pm$ 0.5 & 90.4 $\pm$ 0.5 & 81.2 $\pm$ 0.4 & 85.1 $\pm$ 1.2 \\
MentorMix & 94.0 $\pm$ 0.6 & 94.2 $\pm$ 0.4 & 90.0 $\pm$ 0.7 & 92.0 $\pm$ 0.4 & 84.8 $\pm$ 0.6 & 89.6 $\pm$ 0.8 \\
\bottomrule
\end{tabular}
}
\end{table}

\begin{table}[ht]
\caption{Training with LQM outputs with various techniques. Test accuracy $\pm$ std (\%) on CIFAR10}\label{tab:lqm_plus_tech_cifar10}
\centering
\setlength\tabcolsep{9pt}
% \resizebox{\linewidth}{!}{
\begin{tabular}{c|c|c|c}
\toprule
algorithm	 & \multicolumn{1}{c|}{low (err=0.11)} & \multicolumn{1}{c|}{medium (err=0.19)} & \multicolumn{1}{c}{high (err=0.48)} \\
\midrule
Baseline & 84.1 $\pm$ 0.2 & 78.8 $\pm$ 0.2 & 62.4 $\pm$ 1.1 \\
LQM & 85.6 $\pm$ 0.4 & 81.9 $\pm$ 0.3 & 73.4 $\pm$ 0.3 \\
LQM + Bootstrap & 82.8 $\pm$ 0.9 & 79.8 $\pm$ 0.2 & 73.5 $\pm$ 0.3 \\
LQM + Co-Teaching & 86.3 $\pm$ 0.2 & 80.9 $\pm$ 0.1 & 74.2 $\pm$ 0.4 \\
LQM + MCSoftMax & 85.7 $\pm$ 0.2 & 81.6 $\pm$ 0.1 & 74.2 $\pm$ 0.2 \\
LQM + MentorMix & 86.3 $\pm$ 0.1 & 84.2 $\pm$ 0.3 & 78.4 $\pm$ 0.2 \\
\bottomrule
\end{tabular}
% }
\end{table}

\begin{table}[ht]
\caption{Training with LQM outputs with various techniques. Test accuracy $\pm$ std (\%) on CIFAR100}\label{tab:lqm_plus_tech_cifar100}
\centering
\setlength\tabcolsep{9pt}
% \resizebox{\linewidth}{!}{
\begin{tabular}{c|c|c|c}
\toprule
algorithm	 & \multicolumn{1}{c|}{low (err=0.25)} & \multicolumn{1}{c|}{medium (err=0.38)} & \multicolumn{1}{c}{high (err=0.43)} \\
\midrule
Baseline & 59.2 $\pm$ 1.2 & 52.9 $\pm$ 0.6 & 47.3 $\pm$ 1.0 \\
LQM & 59.4 $\pm$ 1.7 & 53.9 $\pm$ 0.6 & 51.3 $\pm$ 0.9 \\
LQM + Bootstrap & 59.8 $\pm$ 1.1 & 52.8 $\pm$ 0.8 & 49.8 $\pm$ 1.4 \\
LQM + Co-Teaching & 61.0 $\pm$ 1.3 & 56.8 $\pm$ 0.6 & 50.4 $\pm$ 0.7 \\
LQM + MCSoftMax & 63.0 $\pm$ 0.2 & 57.0 $\pm$ 0.4 & 55.2 $\pm$ 0.7 \\
LQM + MentorMix & 62.4 $\pm$ 0.5 & 58.5 $\pm$ 0.2 & 53.7 $\pm$ 0.4 \\
\bottomrule
\end{tabular}
% }
\end{table}

\begin{table}[ht]
\caption{Training with LQM outputs with various techniques. Test accuracy $\pm$ std (\%) on PatchCamelyon}\label{tab:lqm_plus_tech_pcam}
\centering
\setlength\tabcolsep{9pt}
% \resizebox{\linewidth}{!}{
\begin{tabular}{c|c|c|c}
\toprule
algorithm	 & \multicolumn{1}{c|}{low (err=0.1)} & \multicolumn{1}{c|}{medium (err=0.18)} & \multicolumn{1}{c}{high (err=0.23)} \\
\midrule
Baseline & 78.1 $\pm$ 1.8 & 75.1 $\pm$ 0.9 & 72.9 $\pm$ 0.3 \\
LQM & 80.0 $\pm$ 0.1 & 79.1 $\pm$ 0.9 & 77.6 $\pm$ 0.7 \\
LQM + Bootstrap & 80.7 $\pm$ 0.7 & 80.7 $\pm$ 0.9 & 78.2 $\pm$ 1.1 \\
LQM + Co-Teaching & 78.9 $\pm$ 0.1 & 78.2 $\pm$ 1.7 & 76.9 $\pm$ 0.3 \\
LQM + MCSoftMax & 80.3 $\pm$ 0.4 & 80.0 $\pm$ 0.7 & 78.2 $\pm$ 0.9 \\
LQM + MentorMix & 79.5 $\pm$ 0.5 & 77.0 $\pm$ 0.4 & 75.8 $\pm$ 0.2 \\
\bottomrule
\end{tabular}
% }
\end{table}

\begin{table}[ht]
\caption{Training with LQM outputs with various techniques. Test accuracy $\pm$ std (\%) on Cats vs Dogs}\label{tab:lqm_plus_tech_cvd}
\centering
\setlength\tabcolsep{9pt}
% \resizebox{\linewidth}{!}{
\begin{tabular}{c|c|c|c}
\toprule
algorithm	 & \multicolumn{1}{c|}{low (err=0.09)} & \multicolumn{1}{c|}{medium (err=0.20)} & \multicolumn{1}{c}{high (err=0.29)} \\
\midrule
Baseline & 91.9 $\pm$ 0.3 & 87.6 $\pm$ 0.5 & 82.0 $\pm$ 0.5 \\
LQM & 94.9 $\pm$ 0.4 & 92.3 $\pm$ 0.4 & 85.8 $\pm$ 0.4 \\
LQM + Bootstrap & 92.0 $\pm$ 0.4 & 90.5 $\pm$ 0.3 & 89.5 $\pm$ 0.8 \\
LQM + Co-Teaching & 94.1 $\pm$ 0.8 & 91.5 $\pm$ 0.5 & 90.9 $\pm$ 0.5 \\
LQM + MCSoftMax & 93.5 $\pm$ 0.2 & 91.3 $\pm$ 0.2 & 89.1 $\pm$ 0.1 \\
LQM + MentorMix & 94.6 $\pm$ 0.6 & 92.1 $\pm$ 0.5 & 91.1 $\pm$ 0.3 \\
\bottomrule
\end{tabular}
% }
\end{table}

\end{document}